# Performance assessment of the deep learning technologies in grading glaucoma severity

Yi Zhen[1,*], Lei Wang[2,*], Han Liu[2], Jian Zhang[3], Jiantao Pu[2,Δ]

[1]Department of Ophthalmology, Beijing Tongren Hospital, Beijing, China, 10029
[2]Departments of Radiology and Bioengineering, University of Pittsburgh, Pittsburgh, PA, USA, 15213
[3]Department of Ophthalmology, Shaanxi provincial people's hospital, Xi'an, China, 710068
[*]Yi Zhen and Lei Wang contributed equally to this work
[Δ]Corresponding author: Jiantao Pu

## ABSTRACT

**Objective**: To validate and compare the performance of eight available deep learning architectures in grading the severity of glaucoma based on color fundus images.

**Materials and Methods**: We retrospectively collected a dataset of 5978 fundus images and their glaucoma severities were annotated by the consensus of two experienced ophthalmologists. We preprocessed the images to generate global and local regions of interest (ROIs), namely the global field-of-view images and the local disc region images. We then divided the generated images into three independent sub-groups for training, validation, and testing purposes. With the datasets, eight convolutional neural networks (CNNs) (i.e., VGG16, VGG19, ResNet, DenseNet, InceptionV3, InceptionResNet, Xception, and NASNetMobile) were trained separately to grade glaucoma severity, and validated quantitatively using the area under the receiver operating characteristic (ROC) curve and the quadratic kappa score.

**Results**: The CNNs, except VGG16 and VGG19, achieved average kappa scores of 80.36% and 78.22% when trained from scratch on global and local ROIs, and 85.29% and 82.72% when fine-tuned using the pre-trained weights, respectively. VGG16 and VGG19 achieved reasonable accuracy when trained from scratch, but they failed when using pre-trained weights for global and local ROIs. Among these CNNs, the DenseNet had the highest classification accuracy (i.e., 75.50%) based on pre-trained weights when using global ROIs, as compared to 65.50% when using local ROIs.

**Conclusion**: The experiments demonstrated the feasibility of the deep learning technology in grading glaucoma severity. In particular, global field-of-view images contain relatively richer information that may be critical for glaucoma assessment, suggesting that we should use the entire field-of-view of a fundus image for training a deep learning network.

## 1. INTRODUCTION

Glaucoma[1] is a chronic and progressive optic neuropathy caused by the irreversible neurodegeneration of the optic nerve. Glaucoma can gradually reduce the visual field of patients and it ultimately leads to permanent blindness if no proper intervention or treatment is taken. Patients with early glaucoma often have no symptoms of peripheral vision loss, which may last for several years, but could suffer from a severe deterioration of vision with disease progression[2-3]. Currently, there is no cure for this disease. Hence, early screening and precise pathological grading of glaucoma is critical for delaying the disease progression and preventing further damage to the optic nerve. Pathological grading has several potential benefits, such as increasing the efficiency and coverage of screening programs, improving disease diagnosis as a result of early detection and treatment, and reducing the suffering and economic burden of patients. Hence, routine screening is typically regarded as the only way to effectively prevent vision loss.

Conventionally, the detection / diagnosis of glaucoma is performed by trained ophthalmologists in clinical practice, which generally requires an intraocular pressure (IOP) test, a visual field test, and an examination of the optic nerve head[4]. Each of these examinations involves various control parameters, such as gonioscopy and retinal nerve fiber layer (RNFL) evaluations. This process of diagnosing glaucoma is time-consuming[5] and has limited potential in population-based screening for early detection. Moreover, this subjective evaluation is susceptible to intra- and / or inter-observer errors, because clinical experiences play an important role in glaucoma detection and grading[6]. To reduce subjective assessment errors, different clinical indicators or metrics, are designed to aid in the quantitative grading of glaucoma. Among these metrics, the ratio of optic cup to disc, known as the cup-to-disc ratio (CDR)[7-9], is very popular and widely used, as illustrated by the example in Fig. 1, for detecting and classifying glaucoma early, since emerging glaucoma is often associated with increased CDR values[8].

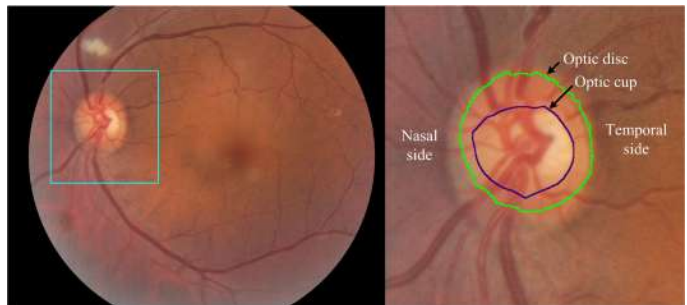

Fig. 1. Illustration of the optic cup and disc (right) depicted on a color fundus image (left).

When assessing glaucoma based on the CDR, a physician often needs to measure the inferior, superior, nasal, and temporal (ISNT) zones[4] in the optic disc area. To gain efficiency and accuracy, investigators have developed a large number of computerized methods[10-13]. Most of these methods focus on segmenting the optic disc and cup from fundus images by leveraging the intensity contrast between the disc region and the surrounding tissues. It has been demonstrated that computerized solutions significantly reduce influencing factors, such as clinical experiences and manual intervention, on glaucoma assessment, and achieve promising accuracy, as compared to the traditional manual procedures[14]. These computerized methods, however, have two main limitations: (1) they are generally based on some hand-crafted image features to identify the boundaries between the optic disc and cup, which is not easy and needs a wide variety of background knowledge and information, such as computer vision, image processing, and clinical information; (2) only a small amount of image information surrounding the optic disc is typically utilized in these methods without considering other valuable information away from the disc region (e.g., small blood vessels and fovea). This means that a great deal of texture information in regard to the retina is ignored. These limitations may be the underlying reason of performance bottlenecks for the traditional approaches when used for grading glaucoma.

Nowadays, deep learning technology[15-17], known as convolutional neural networks (CNNs), has gained considerable popularity in the field of computer vision. Deep learning technology is an efficient way to detect objects and classify images, as compared to traditional image processing methods, because of its unique capability of implicitly extracting discriminative features in an image and optimally combing these features for classification purposes[18]. Unfortunately, very limited effort has been made up to this point to assess glaucoma severity using deep learning technology. In addition, deep learning often needs large datasets with labeled information[19-21] and has special requirements for computer hardware in both memory and computation capability. Given retinal fundus images, which often have relatively high image resolution[22] (e.g., 2500×2500 pixels), it is impossible at this time for the most state-of-art Graphics Processing Unit (GPU) to fully utilize this high resolution characteristic. Hence, it is necessary to assess whether the deep learning technology[23-25] has better performance on global images with low resolution or local images with high resolution in grading glaucomatous optic neuropathy.

In this study, we collected a relatively large dataset to investigate whether the glaucoma severity can be accurately assessed using the deep learning technology on the basis of color fundus images. Several classical deep learning architectures were applied to the collected dataset consisting of 5978 fundus images acquired on different subjects. The glaucoma severities were labeled by the consensus of two ophthalmologists. We specifically investigated the image resolution on the classification performance by generating two sets of images in terms of resolution. One set is a relatively low resolution by resizing the entire images to smaller ones, and the other one is the relatively high resolution by cropping the local disc regions from the original images since CDR is widely applied for the detection and staging of glaucoma. A detailed description of the involved methods and the experimental results follows.

## 2. MATERIALS AND METHODS

### 2.1. Image datasets

To train and validate several available CNN architectures, we retrospectively collected a fundus image dataset acquired on different subjects in the Beijing Tongren Hospital using fundus cameras from multiple manufacturers (i.e., Canon and Topcon). These images have a resolution ranging from 1924×1556 pixels to 2336×2336 pixels. An honest broker de-identified the images and two experienced ophthalmologists (>10 years' experience) independently rated the images using a four-scale strategy, namely 0, 1, 2 or 3 for none, mild, moderate, or severe glaucoma, respectively. We used the ophthalmologists' consensus grading as the ground truth, and ignored those images that they disagreed regarding the classification. We obtained a total of 5978 images for this study as summarized in Table 1. We randomly divided each class of the image samples into three entirely independent sub-groups for training, validation, and testing purposes. For the testing dataset, we made the number of images in each category the same to eliminate potential bias caused by an imbalance in sample size.

Table 1: The collected fundus images used for glaucoma classification according to their severity.

| Stage | None | Mild | Moderate | Severe |
|---|---|---|---|---|
| Training | 1534 | 947 | 1224 | 757 |
| Validation | 384 | 237 | 306 | 189 |
| Test | 100 | 100 | 100 | 100 |
| Total | 2018 | 1284 | 1630 | 1046 |

### 2.2. Image pre-processing

We firstly processed the collected fundus images to exclude unnecessary margins / backgrounds that do not contain any useful information. Only the field-of-view of an image was preserved. Then, we resized the cropped images to the same dimension and normalized using a histogram equalization to lower the influence caused by different illumination conditions and resolutions. Considering that CDR is commonly used for the detection and staging of glaucoma, we further cropped the disc regions from the field-of-view images, as shown in Fig. 2. The cropped local region had a margin that was approximately equal to the diameter of the optic disc to assure that surrounding regions were included. This not only retains a considerable amount of discriminative feature information closely related with glaucoma, but also significantly excludes "irrelevant" information, which accounts for a large part of fundus images, to improve computational efficiency. Since the optic disc appears as a circular region, we used a circular ROI in this study.

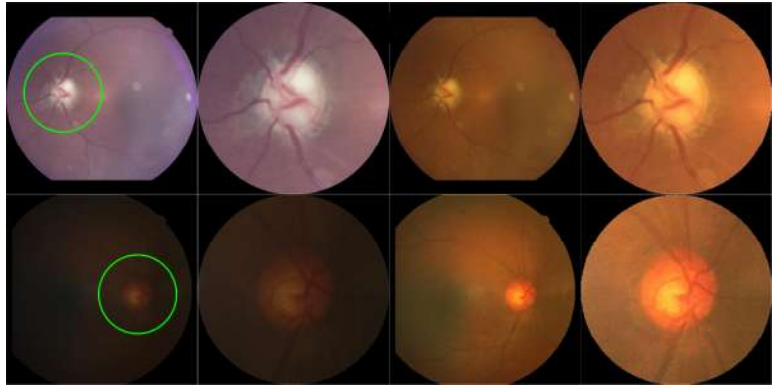

Fig. 2. Illustration of the global field-of-view images and the disc images. The first two columns are the original fundus images and their local regions surrounding the optic disc, the last two columns are their preprocessed versions, respectively.

**2.3 Deep learning architectures**

Deep convolutional neural networks[15] have been developed and applied for a wide variety of image classification applications. Their strength lies in weight sharing, based on the assumption that an image often contains many similar structures in different locations. These structures can be detected by a set of convolution filters with different kernels, followed by an element-wise nonlinear activation transform. Convolutional results are alternated with the pooling operation to reduce the dimension of feature information and to induce a certain amount of translation invariance. After a set of convolutional and pooling processes, fully connected (FC) layers are used for classification purposes. Following the similar procedures, eight representative CNN architectures[26-30] (termed as VGG16, VGG19, ResNet, DenseNet, InceptionV3, InceptionResNet, Xception, and NASNetMobile, respectively) were developed for the ILSVRC challenge[17]. The CNN architectures became popular due to their broad applicability in many image processing tasks. In this study, we applied these classical architectures to the collected images to investigate whether the high-resolution disc images are superior to the global field-of-view images in assessing glaucoma.

VGG16 and VGG19[16] were developed to explore the influence of the network depth on classification accuracy. They solely employ a number of 3×3 convolution kernels and the Rectifier Linear Unit (ReLU) activation function for each convolutional layer, and a 2×2 pixel window for each max-pooling layer. A simple stack of convolutional and pooling layers is followed by three FC layers with 4096, 4096, and 1000 channels, respectively. The only difference between the two architectures is the number of convolution layers concatenated between two adjacent max-pooling layers, which leads to varying receptive fields for detecting similar structures. Comparing VGG16 and VGG19 reveals that deeper networks typically have better classification performance. Unlike VGG16 and VGG19, ResNet and DenseNet[23-24] apply a number of 1×1 and 3×3 convolution kernels, together with the batch normalization (BN) scheme[26], to construct different convolution blocks. These blocks are alternatively stacked to perform image classification, thus causing a significant increase in the network depth. In these convolution blocks, the 1×1 convolution kernels are mainly used to reduce the amount of image information, while batch normalization can improve classification accuracy. Based on the number of stacked blocks, ResNet and DenseNet have different depths and performances. Both of them tend to consistently improve accuracy as network depth increases, since they are able to effectively alleviate the problem of vanishing or exploding gradients with increasing network depth. In this study, we used ResNet and DenseNet with depths of 50 and 121, respectively, for classifying glaucoma. Apart from above four networks, InceptionV3, InceptionResNet, and Xception are three popular variants of the GoogLeNet architectrue[27-29] by defining different inception blocks. These blocks are critical for classification performances, and regarded as a filter constructed by several sub-filters of various dimensions and sizes. In each convolution block, the input is branched into different convolutional sub-networks, and the outputs of all sub-networks are concatenated as the output of this block. These blocks have the potential for capturing various discriminative features in different hierarchies and positions. They do not simply stack 3×3, 5×5 or 7×7 convolution kernels to process the input, and hence avoid high computational cost. NASNetMobile[30] is one of the variants of NASNet architecture for mobile platforms. It employed the Neural Architecture Search (NAS) method to search for the best convolutional layers (or "cell") on relatively small image datasets. The convolutional cells were then used for the dataset of interest for better classification performances and smaller computational budgets. With these convolutional cells, Normal and Reduction cells were constructed to expand the application of NASNet for images with any size. The former extracted a set of feature maps of the input images, without changing image dimension, while the latter returned a feature map whose height and width were reduced by a factor of two. By combining these Normal and Reduction cells in different ways, there was a family of NASNet architectures that achieved accuracies superior to most available CNN architectures at equivalent or smaller computational budgets.

### 2.4 Training schemes

We trained the above CNN architectures to classify glaucoma in terms of severity using the global and local image sub-groups. We then fine-tuned these models in order to take full advantage of pre-trained weights by adding two FC layers and one dropout regularization for the first FC layers with a probability of 0.5. These two FC layers had channels of 1024 and 4, respectively, where the last FC layer, activated by the softmax function, represented the prediction probabilities of glaucomatous image ROIs for each class. For comparison purposes, we resized both global and local ROIs to 224×224, which enabled us to use available classification configurations from the ILSVRC challenge. We optimized the convolution parameters in each CNN architecture to minimize the cross entropy loss[31] using stochastic gradient descent with a batch size of 32 samples, a momentum of 0.95, and an initial learning rate of 0.001. The rate was decreased by a factor of 10 for every 30 epochs. The number of training epochs was assigned to 150, with a division point of 0.5 for prediction probabilities. To prevent overfitting, we employed data augmentation techniques[32] to artificially enlarge the training set via some label-preserving image transformations. Specifically, each ROI was randomly transformed using horizontal and vertical flip in training, and rescaled with a factor ranging from 0.85 to 1.15. We also applied image rotation[33] along each axis by the values of -30~30 in degree.

### 2.5 Performance assessment

Both the original and fine-tuned CNN architectures mentioned above were constructed through the Keras library[34]. Classification experiments were performed on a PC with a 3.50GHz Intel(R) Xeon(R) E5-1620 CPU, 8GB RAM and NVIDIA Quadro P5000. The classification performance was assessed quantitatively using the precision, recall, F1 score, and quadratic weighted kappa, respectively:

$$P = \frac{Tp}{Tp + Fp} \quad (1)$$

$$R = \frac{Tp}{Tp + Fn} \quad (2)$$

$$F1 = \frac{2PR}{P + R} \quad (3)$$

$$\kappa = 1 - \frac{\sum_{ij} \omega_{ij} O_{ij}}{\sum_{ij} \omega_{ij} E_{ij}} \quad (4)$$

$$\omega_{ij} = \frac{(i - j)^2}{(N - 1)^2} \quad (5)$$

where $Tp$, $Fp$, and $Fn$ denote true positive (belonging to a given class $L$ and correctly classified), false positive (not belonging to a given class $L$ and incorrectly classified as $L$) and false negative (belonging to a given class $L$ and incorrectly classified), respectively. $N$ is the number of classes. $i$ and $j$ are the class labels assigned by an ophthalmologist and method for a given image, respectively. The weight $\omega_{ij}$ is calculated based on the difference between labels $i$ and $j$. Matrix $O_{ij}$ is the number of images classified simultaneously as $i$ by the ophthalmologist and $j$ by the method. Matrix $E_{ij}$ is the outer product between the histogram vector of the clinician and method. The quadratic weighted kappa metric[35] is a measure of inter-rater agreement between two raters. The quadratic weighted kappa metric is robust and accurate for multi-class classifications, and thus was widely used in recent deep learning contests hosted on Kaggle[36-38]. This metric often ranges from 0 (random agreement) to 1 (complete agreement). In addition, the area under the receiver operating characteristic (ROC) curve (AUC) was computed using scikit-learn python package. The AUC is the probability of the true positive assessment as a function of the probability of the false positive assessment. The value of the AUC is 1 for perfect classification and 0.5 for random guessing. A 95% confidence interval (CI) is used for both AUC and the quadratic weighted kappa.

## 3. RESULTS

Tables 2 and 3 summarize the classification results when validating the eight CNN architectures trained from scratch on the global and local ROIs on the testing sets. These architectures achieved average quadratic weighted kappa scores of 80.51% (95% CI, 80.38%-80.61%) and 78.83% (95% CI, 78.68%-78.93%), and average total classification accuracies (i.e., the ratio between correctly classified image samples and the whole samples) of 65.66% and 62.73% for the global field-of-view images and the local disc images, respectively. Among these CNN architectures, VGG19, DenseNet, InceptionV3, Xception and NASNetMobile had significantly improved quadratic weighted kappa scores (i.e., 1.24%, 4.76%, 4.96%, 6.7% and 8.92%, respectively) when trained on global ROIs, as compared to their counterparts trained on local ROIs. VGG16, ResNet, and InceptionResNet demonstrated slightly better capabilities of classifying the local disc images (i.e., 0.66%, 8.92% and 3.56%, respectively). Hence, the local disc images were, on average, inferior to the global field-of-view images for glaucoma classification.

**Table 2**. Performances of eight CNN architectures trained from scratch on global images using the precision, recall, F1 score and quadratic weighted kappa (%), respectively.

| CNNs | Precision | | | | Recall | | | | F1 score | | | | kappa |
|---|---|---|---|---|---|---|---|---|---|---|---|---|---|
| | 0 | 1 | 2 | 3 | 0 | 1 | 2 | 3 | 0 | 1 | 2 | 3 | |
| VGG16 | 94.34 | 72.31 | 37.58 | 59.52 | 89.29 | 47.00 | 59.00 | 50.00 | 91.75 | 56.97 | 45.91 | 54.35 | 80.56 |
| VGG19 | 91.74 | 72.04 | 42.74 | 65.12 | 89.29 | 67.00 | 53.00 | 56.00 | 90.50 | 69.43 | 47.32 | 60.22 | 81.36 |
| ResNet | 87.72 | 51.28 | 34.52 | 43.96 | 89.29 | 20.00 | 58.00 | 40.00 | 88.50 | 28.78 | 43.28 | 41.89 | 69.26 |
| DenseNet | 95.24 | 67.53 | 41.96 | 67.82 | 89.29 | 52.00 | 60.00 | 59.00 | 92.17 | 58.76 | 49.38 | 63.10 | 83.20 |
| InceptionV3 | 96.08 | 64.76 | 42.59 | 67.74 | 90.74 | 68.00 | 46.00 | 63.00 | 93.33 | 66.34 | 44.23 | 65.28 | 83.71 |
| InceptionResNet | 93.07 | 64.13 | 37.35 | 55.30 | 90.38 | 59.00 | 31.00 | 70.19 | 91.71 | 61.46 | 33.88 | 61.86 | 78.83 |
| Xception | 94.23 | 71.43 | 42.33 | 72.00 | 94.23 | 50.00 | 69.00 | 50.00 | 94.23 | 58.82 | 52.47 | 59.02 | 85.03 |
| NASNetMobile | 90.38 | 59.56 | 44.00 | 80.56 | 90.38 | 77.88 | 42.31 | 58.00 | 90.38 | 67.50 | 43.14 | 67.44 | 82.11 |

**Table 3**. Performances of eight CNN architectures trained from scratch on local images using the precision, recall, F1 score and quadratic weighted kappa (%), respectively.

| CNNs | Precision | | | | Recall | | | | F1 score | | | | kappa |
|---|---|---|---|---|---|---|---|---|---|---|---|---|---|
| | 0 | 1 | 2 | 3 | 0 | 1 | 2 | 3 | 0 | 1 | 2 | 3 | |
| VGG16 | 89.33 | 54.21 | 49.09 | 80.00 | 72.04 | 62.37 | 58.06 | 68.82 | 79.76 | 58.00 | 53.20 | 73.99 | 81.22 |
| VGG19 | 86.49 | 51.96 | 48.78 | 79.01 | 68.82 | 56.99 | 64.52 | 63.37 | 76.65 | 54.36 | 55.56 | 70.33 | 80.12 |
| ResNet | 93.33 | 49.09 | 48.39 | 76.83 | 57.73 | 58.06 | 64.52 | 67.74 | 71.34 | 53.20 | 55.30 | 72.00 | 78.18 |
| DenseNet | 84.42 | 45.65 | 39.37 | 77.38 | 69.89 | 45.16 | 53.76 | 64.36 | 76.47 | 45.40 | 45.45 | 70.27 | 78.44 |
| InceptionV3 | 89.47 | 53.66 | 46.32 | 73.17 | 70.10 | 47.31 | 67.74 | 64.52 | 78.61 | 50.29 | 55.02 | 68.57 | 78.75 |
| InceptionResNet | 85.37 | 57.14 | 48.51 | 80.52 | 72.16 | 55.91 | 67.01 | 63.92 | 78.21 | 56.52 | 56.28 | 71.27 | 82.39 |
| Xception | 87.01 | 50.00 | 41.46 | 73.61 | 69.07 | 58.06 | 52.58 | 56.99 | 77.01 | 53.73 | 46.36 | 64.24 | 78.33 |
| NASNetMobile | 77.91 | 46.55 | 42.68 | 80.88 | 72.04 | 29.03 | 72.16 | 59.14 | 74.86 | 35.76 | 53.64 | 68.32 | 73.19 |

Tables 4 and 5 show the classification results when applying the fine-tuned CNN architectures trained using pre-trained weights to the testing images. Our experiments demonstrated that the fine-tuned architectures had higher classification accuracy when trained on the global field-of-view images, as compared to the counterparts trained on the local disc images. Specifically, VGG16 and VGG19 were trapped into local minima and incompetent for this classification task. This suggests they have relatively limited classification performances among these architectures. Different from VGG16 and VGG19, the other CNN architectures obtained average quadratic weighted kappa scores of 85.29% (95% CI, 85.17%-85.36%) and 82.72% (95% CI, 82.63%-82.73%) for the global field-of-view images and the local disc images, respectively. Their total classification accuracy based on the global ROIs increased by 5.29% on average than their counterparts trained on the local ROIs. Additionally, DenseNet and InceptionResNet had the highest classification accuracy of 75.50% for global ROIs.

Table 4. Performances of fine-tuned CNN architectures trained using the pre-trained imagenet weights on global images in terms of the precision, recall, F1 score and quadratic weighted kappa (%), respectively.

| CNNs | Precision | | | | Recall | | | | F1 score | | | | kappa |
|---|---|---|---|---|---|---|---|---|---|---|---|---|---|
| | 0 | 1 | 2 | 3 | 0 | 1 | 2 | 3 | 0 | 1 | 2 | 3 | |
| VGG16 | 0.00 | 0.00 | 25.00 | 0.00 | 0.00 | 0.00 | 89.29 | 0.00 | 0.00 | 0.00 | 39.06 | 0.00 | 0.00 |
| VGG19 | 25.00 | 0.00 | 0.00 | 0.00 | 89.29 | 0.00 | 0.00 | 0.00 | 39.06 | 0.00 | 0.00 | 0.00 | 0.00 |
| ResNet | 92.59 | 72.83 | 48.09 | 75.29 | 89.29 | 67.00 | 63.00 | 61.54 | 90.91 | 69.79 | 54.54 | 67.72 | 85.38 |
| DenseNet | 94.29 | 75.86 | 51.20 | 73.74 | 91.67 | 66.00 | 64.00 | 67.59 | 92.96 | 70.59 | 56.89 | 70.53 | 88.91 |
| InceptionV3 | 94.34 | 63.96 | 44.53 | 77.46 | 89.29 | 71.00 | 57.00 | 52.88 | 91.75 | 67.30 | 50.00 | 62.85 | 84.36 |
| InceptionResNet | 93.46 | 67.83 | 51.55 | 76.29 | 89.29 | 78.00 | 50.00 | 71.15 | 91.33 | 72.56 | 50.76 | 73.63 | 88.91 |
| Xception | 92.45 | 67.47 | 39.53 | 76.60 | 90.74 | 56.00 | 68.00 | 36.00 | 91.59 | 61.20 | 50.00 | 48.98 | 79.96 |
| NASNetMobile | 97.03 | 68.66 | 43.79 | 70.33 | 90.74 | 46.00 | 67.00 | 61.54 | 93.78 | 55.09 | 52.96 | 65.64 | 84.22 |

Table 5. Performances of fine-tuned CNN architectures trained using the pre-trained imagenet weights on local images in terms of the precision, recall, F1 score and quadratic weighted kappa (%), respectively.

| CNNs | Precision | | | | Recall | | | | F1 score | | | | kappa |
|---|---|---|---|---|---|---|---|---|---|---|---|---|---|
| | 0 | 1 | 2 | 3 | 0 | 1 | 2 | 3 | 0 | 1 | 2 | 3 | |
| VGG16 | 0.00 | 0.00 | 0.00 | 25.00 | 0.00 | 0.00 | 0.00 | 89.29 | 0.00 | 0.00 | 0.00 | 39.06 | 0.00 |
| VGG19 | 25.00 | 0.00 | 0.00 | 0.00 | 89.29 | 0.00 | 0.00 | 0.00 | 39.06 | 0.00 | 0.00 | 0.00 | 0.00 |
| ResNet | 81.11 | 59.46 | 49.01 | 80.82 | 75.26 | 45.36 | 76.29 | 60.82 | 78.08 | 51.46 | 59.68 | 69.41 | 83.92 |
| DenseNet | 87.18 | 54.41 | 47.66 | 73.58 | 70.10 | 39.78 | 65.59 | 80.41 | 77.71 | 45.96 | 55.21 | 76.84 | 82.64 |
| InceptionV3 | 84.27 | 60.20 | 50.91 | 78.16 | 74.26 | 63.44 | 60.22 | 70.10 | 78.95 | 61.78 | 55.18 | 73.91 | 85.57 |
| InceptionResNet | 80.68 | 56.32 | 50.91 | 75.76 | 73.20 | 50.52 | 60.22 | 77.32 | 76.76 | 53.26 | 55.18 | 76.53 | 84.36 |
| Xception | 90.79 | 52.70 | 48.34 | 80.72 | 71.13 | 40.21 | 78.49 | 69.07 | 79.77 | 45.62 | 59.83 | 74.44 | 82.91 |
| NASNetMobile | 79.80 | 52.33 | 40.58 | 73.77 | 78.22 | 46.39 | 60.22 | 48.39 | 79.00 | 49.18 | 48.49 | 58.44 | 76.91 |

Fig. 3 shows the ROC curves of the original and fine-tuned CNN architectures to demonstrate their performance differences, though total classification accuracy was more objective and intuitive. The classification results of these CNN architectures (except VGG16 and VGG19) and their fine-tuned versions achieved average AUCs of 90.00% (95% CI, 89.93%-90.07%) and 92.67% (95% CI, 92.61%-92.73%) when trained on the global field-of-images using the different training procedures, respectively. These CNN architectures obtained average AUCs of 87.00% (95% CI, 86.92%-87.07%) and 90.00% (95% CI, 89.94%-90.06%) when using the local disc images, without changing any other experiment configurations, suggesting that the image regions far away from the optic disc actually played an important role, or have some unique texture information, in glaucoma assessment. In addition, the original architectures, except VGG16 and VGG19, had, on average, inferior performances than the fine-tuned versions, regardless of the global or local images. Fig. 4 shows in more detail the classification performance of the original DenseNet architecture and its fine-tuned version using the confusion matrix. As demonstrated by these matrixs, most errors appeared close to the diagonal line. This demonstrates that most misclassifications happened in the adjacent classes, which often occurs for glaucoma classification by an ophthalmologist. Specifically, 97.50% and 99.00% of the testing cases had less-than-one neighboring class errors when trained from scratch and using pre-trained weights on global field-of-view images, respectively. These errors were, to some extent, acceptable for certain cases, which may be further reduced by post-processing algorithms[39-40] to suppress the gross errors. These matrixs also display that images with mild and moderate glaucoma cannot be easily differentiated, since these categories shared lots of key image features.

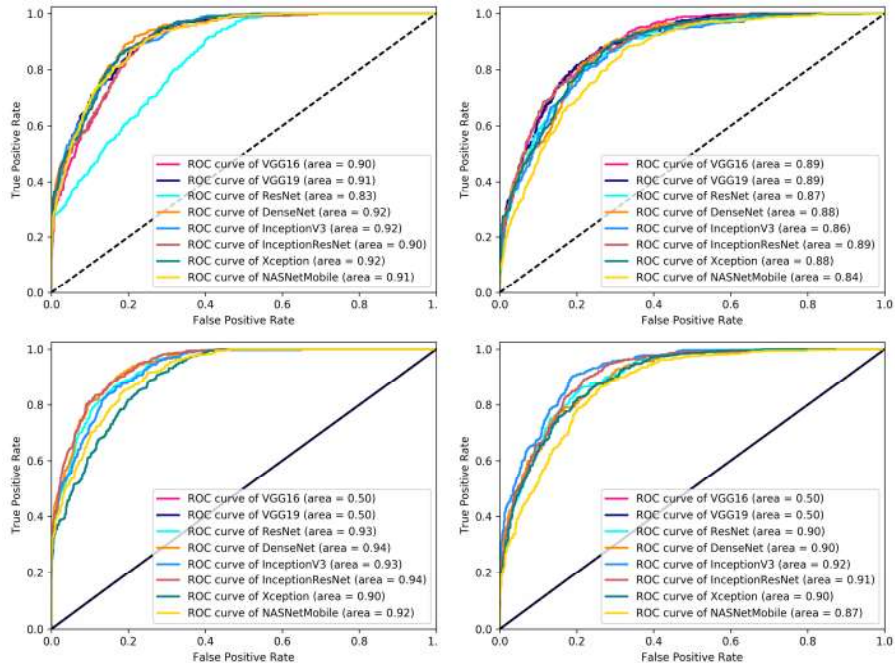

Fig. 3. The ROC curves of all CNN architectures. The first row is the result of original architectures trained from scratch on global field-of-view images and the local disc images, respectively; the second row is their fine-tuned results trained using the pre-trained weights.

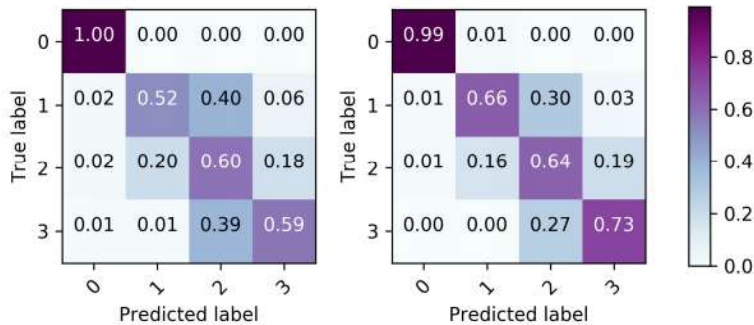

Fig. 4. Confusion matrixs of the original and fine-tuned DenseNet for global field-of-view images, respectively.

## 4. DISCUSSION

It is not a trivial task to accurately grade glaucoma in clinical practices, even for experienced ophthalmologists, due to the underlying complex image textures associated with glaucoma development. Ophthalmologists have moved a long way in the past couple of decades to come to an agreement that optic disc cupping – particularly cup-to-disc ratio (CDR) – is an error-prone way to stage glaucoma[41-42]. This is due not only to high intra- and inter-grader variability using disc photos but most importantly because the optic disc morphology is highly variable (including the effect of disc size) and estimates of cupping or CDR hold a weak relationship with glaucoma severity as assessed with visual fields. This suggests that the estimation of CDR has limited capability of assessing glaucoma severity based solely on disc photos. However, disc photos are still used largely, as a widely available and inexpensive tool, for initial assessment of glaucomatous optic neuropathy in many places, even though assessment accuracy is relatively low.

The emerging deep learning technology may be a good choice for this specific task due to its unique capability of learning the image textures / features. When utilizing the deep learning technology, we need to significantly reduce the resolution of the original fundus images due to the limits of computer hardware in memory and computation. Whether the high-resolution regional disc images can enable a higher assessment performance as compared to the low-resolution field-of-view images is not clear in glaucoma classification. We believe that clarification of this specific issue as addressed in this study is very important for applying the deep learning technology to the automated assessment of glaucoma severity. In this study, we compared and validated the performance of several classical deep learning technologies in grading the severity of glaucomatous optic neuropathy and in particular investigated the influence of the global field-of-view images and the local disc images in classification performance. We note that the emphasis of this study is not on the development of novel deep learning architectures. Our experimental results demonstrated the unique strength of the deep learning in assessing glaucoma severity and the global ROIs were superior to the local ROIs in terms of classification accuracy, though image resolution had been significantly reduced. This resolution reduction did not completely prevent the detection of certain important features associated with glaucoma.

As compared with available conventional machine learning based approaches[43], which had AUCs ranging from 66.7% to 79.0%, the CNN architectures demonstrated very promising performance in grading the severity of glaucoma with an AUC of 94.0%. By segmenting and quantifying the optic disc and cup, Cheng et al.[11] reported a higher performance with an AUC of 80.0% and 82.2% for two separate datasets. Although the datasets used for the development and validation in these studies are different, the statistical assessment suggests a relatively higher performance of the CNN architectures. It is notable that up to this point very limited effort has been performed to grade or classify glaucoma using the deep learning technology. Cerentini et al.[44] trained the GoogLeNet architecture on 787 fundus images to identify glaucoma and obtained 87.6% detection accuracy. Chen et al.[45] proposed a CNN architecture for glaucoma screening and achieved an average AUC of 85.90%. Orlando et al.[46] differentiated normal and glaucomatous images based on a deep learning approach, with an AUC of 76.26%. These available CNN-based investigations primarily focused on classifying the fundus images into two simple categories, namely glaucoma and non-glaucoma, and their datasets were relatively small. In contrast, we collected a relatively large dataset and classified the fundus images into four categories with a quadratic weighted kappa score of 88.91% and an AUC of 94.0%.

Comprehensive performance comparison of these typical CNN architectures for glaucoma classification is not only helpful for our understanding of their unique characteristics, but also may aid in developing novel CNN frameworks for this specific issue. As demonstrated by our experiments, VGG16 and VGG19 had relatively low classification accuracy and robustness by simply stacking 3×3 convolution kernels. This stack leads to a rapid increase in convolution parameters, and potentially limits the receptive fields for capturing certain important features. In contrast, the other architectures achieved reasonable accuracy with different well-handcrafted convolution blocks. These blocks had sophisticated neural network structures and depths, relative to the plain 3×3 convolution kernel. Such a design is capable of detecting a large number of discriminative features with varying hierarchies and receptive fields. In particular, the block in the DenseNet had better capability for feature detection, suggesting that the dense integration of non-adjacent convolutional layers were helpful for assessing glaucoma.

We are aware of some limitations in this study. First, although we collected a relatively larger dataset for grading glaucoma, as compared with available studies[42-44], this dataset is actually small for deep learning, especially when considering the variation of the collected images in intensity, quality, and other acquisition protocols. Consequently, the trained architectures may limit the capability of the CNN models to differentiate adjacent categories. For instance, Kermany et al.[20] and Gulshan et al.[19] used more than 100,000 images in their studies. Second, the manual assessment / labeling of the collected image dataset may not be perfect. Although the consensus of two experienced ophthalmologists were used as a way to alleviate the potential bias of subjective assessment, it is difficult to accurately grade two adjacent categories, because fundus image based diagnosis of glaucoma is particularly difficult and heavily depends on clinical experiences. This underlying bias may affect the performance of the deep learning network. In a previous study performed by Gulshan et al.[19], fifty-four experts were employed to grade diabetic retinopathy. Third, as we explained, our primary emphasis is not on the development of novel deep learning architectures for grading glaucoma severity in this study, but the findings in this study may be helpful for the future development of novel / optimal deep learning based methods for glaucoma assessment.

## 5. CONCLUSION

We performed a comparative investigation of several classical deep learning architectures in assessing glaucoma severity and in particular investigate whether local disc images are superior to global field-of-view images. A dataset consisting of 5978 images was collected for training and validation purposes. Our experimental results demonstrated the very promising potential of deep convolutional neural networks for glaucoma assessment and, in particular, showed that global field-of-view images contain important information associated with glaucoma despite their relatively low resolutions.

## Acknowledgements

This work is supported by National Institutes of Health (NIH) (Grant No. R21CA197493, R01HL096613) and Jiangsu Natural Science Foundation (Grant No. BK20170391).